# Three-layer deep learning network random trees for fault detection in chemical production process

Ming Lu, Zhen Gao, Ying Zou, Zuguo Chen, Pei Li

School of Information and Electrical Engineering, Hunan University of Science and Technology, Xiangtan, China

**Correspondence**

Zhen Gao, School of Information and Electrical Engineering, Hunan University of Science and Technology, Xiangtan, China.

Email: Z.Gao@mail.hnust.edu.cn

**Funding information**

National Natural Science Foundation of China, Grant/Award Numbers: 62203164, 62373144; Scientific Research Fund of Hunan Provincial Education Department (Outstanding Young Project), Grant/Award Number: 21B0499; Hunan Provincial Department of Education, Grant/Award Number: 22A0349

**Abstract**

With the development of technology, the chemical production process is becoming increasingly complex and large-scale, making fault detection particularly important. However, current detection methods struggle to address the complexities of large-scale production processes. In this paper, we integrate the strengths of deep learning and machine learning technologies, combining the advantages of bidirectional long and short-term memory neural networks, fully connected neural networks, and the extra trees algorithm to propose a novel fault detection model named three-layer deep learning network random trees (TDLN-trees). First, the deep learning component extracts temporal features from industrial data, combining and transforming them into a higher-level data representation. Second, the machine learning component processes and classifies the features extracted in the first step. An experimental analysis based on the Tennessee Eastman process verifies the superiority of the proposed method.

# 1 INTRODUCTION

In the rapidly developing chemical production field, the emergence of intelligent control systems presents new opportunities and challenges[1]. The control system can intelligently schedule and utilize resources to maximize production efficiency. However, the complex processes and harsh operating conditions in chemical production expose control systems to risks like toxic corrosion and safety management challenges[2]. Interdependence of system components means that failure of any component can cause cascading failures, leading to serious property damage[3]. This risk is compounded by the production process's instability. These issues underline the urgent need for intelligent and effective fault detection and diagnosis (FDD) methods in the chemical industry to make the production process more flexible and controllable[4,5].

The development and refinement of FDD methods have long been a key research focus. Traditional FDD methods include nonlinear observer-based method, filter-based method, differential geometry method, and so on. The nonlinear observer-based method converts the nonlinear fault detection problem into a linear fault detection problem for special nonlinear systems[6]; The filter-based method generates residuals at the equilibrium point of the system for local linearization for nonlinear discrete systems[7]; The differential geometry method decomposes the system in state transformations, and designs observers for the decomposed subsystems to realize the detection and separation of faults[8]. However, with the advent of the big data era, the complex and huge amount of data makes it difficult for these methods to maintain accuracy and timeliness in FDD[9].

The above shifts have promoted the adoption of FDD methods based on data-driven approaches, which are no longer based on traditional physical models or theoretical knowledge for troubleshooting, but rather on understanding the system behaviour by analyzing large amounts of data[10]. Data-driven FDD methods include statistical-based methods, machine learning-based methods, and deep learning-based methods. Statistical-based methods such as principal component analysis (PCA) and partial least squares (PLS). Both PCA and PLS belong to the data dimensionality reduction techniques. PCA simplifies data structure by transforming it into linearly uncorrelated variables through orthogonal transformation[11]. Alakent et al. proposed the $ICA_{pIso}$-PCA method to construct a nonlinear fractional matrix, and apply the ICA-PCA method to realize the detection and isolation of nonlinear faults[12]. It addresses the issue of traditional methods which are highly susceptible to the smearing effect. However, the method currently does not fully account for the dynamic characteristics in measurement matrix. PLS builds linear regression models between multiple predictor and response variables to elucidate industrial processes[13]. However, neither can be applied to nonlinear systems. To address this issue, kernel principal component analysis (KPCA) was developed. KPCA is the nonlinear extension of PCA, which maps the original features to a high-dimensional space via the kernel function to make the nonlinear structure linearly separable, and then performs PCA in the space[14]. KPCA can address the nonlinear challenges in complex industrial processes. However, selecting the suitable KPCA kernel and adjusting its parameters relies on prior experience.

Machine learning-based FDD methods such as the support vector machine (SVM), Manifold Learning-based method and random forest algorithm (RF), etc[15]. SVM is suitable for nonlinear, high-dimensional systems, constructing hyperplanes to separate variable state classes in a multidimensional space for fault detection[16,17]. However, SVM is a binary classification algorithm, requiring an extension strategy for multiclassification problems, which causes an additional computational load. Manifold Learning-based method aims to capture low-dimensional, embedded representation of high-dimensional data. Zhang et al. proposed a manifold-based data monitoring method by integrating distance and angle information between point pairs to address the issues of inaccurate downscaling in high-dimensional data and the underutilization of information[18]. However, the method is designed for manufacturing process data under a single working condition, with limited effect in multiple conditions. RF is an integrated learning algorithm, which realizes fault detection by constructing a decision tree with multiple subsets of different features[19]. RF performs well on multiclassification problems but is not an optimal choice for high-dimensional data or structured data. The Extra Trees algorithm (ET) improves upon RF by introducing greater randomness. Arya M et al. utilized ET to select the best subset of features, which were fed into a deep learning network for early-stage diabetes prediction with an accuracy of over 97%[20]. S. Yousefi et al. applied ET to machine part fault detection, optimizing ET parameters with the Bayesian optimization method, reaching up to 99% accuracy[21].

As industrial processes become more automated and complex. Deep learning-based FDD methods are becoming the cutting edge of the field[22]. Deep learning-based FDD methods include autoencoder (AE), convolutional neural network (CNN), and long-short-term memory network (LSTM), etc. AE can learn the compressed representation of input data and reconstruct input data, then determine the occurrence of faults by monitoring the reconstruction errors of fault data[23]. However, the feature extraction and recognition ability of AE are weaker than those of some supervised learning algorithms. CNN performs well in the domains with spatial correlation such as image recognition and video analytics, it identifies the fault patterns by learning the spatial hierarchies in the data to realize fault detection[24]. Yuan et al. proposed a method named variable correlation analysis-based convolutional neural network (VCA-CNN) for far topological feature extraction and industrial predictive modelling, and was validated on hydrocracking and debutanizer column process[25]. However, its special convolution kernel is mainly used for CNN and has limitations in generalizing to other models. LSTM, ideal for data with temporal correlations, learns long and short-term dependencies through a gating mechanism to capture anomalous patterns over time[26]. Zhang S et al. innovatively combined LSTM with a trapezoidal auto-encoder for application to continuous stirred-tank heaters and the Tennessee Eastman Benchmark Process, achieving over 95% fault detection rate[27]. Because the number of hidden layer nodes in LSTM has impacts on the accuracy of fault detection and the number of iterations to find the optimal solution, Han Y et al. determined the optimal number of LSTM hidden layer nodes for various faults by comparing training errors, resulting in enhanced fault detection accuracy[28]. Yuan et al. proposed a method named Attention-Based Interval Aided Networks (AIA-Net) for modelling multivariate time-series data with heterogeneous sample intervals and missing values, and is successfully applied to predict the C5 and C6 content in the light naphtha during real hydrocracking process[29].

To address the above background, we integrate the strengths of deep learning techniques and machine learning techniques, combine the advantages of bidirectional long and short-term memory neural network (BLSTM), LSTM, fully connected neural network (FCNN), and ET, and propose a new fault detection model named three-layer deep learning network random trees (TDLN-trees). TDLN-trees starts with the sliding window method to extract offline samples of industrial process variables, which ensures the temporal integrity and relevance of the data. Next, the fault data are normalized with normal operating state data, and the fault labels are one-hot encoded. In the process of online fault detection, TDLN-trees uses the multiple LSTM and FCNN structures in its Deep Learning component (DL) to capture and fit complex features from the time series data, then ET in the Machine Learning component (ML) to realize fault classification.

In the experimental section, the proposed method is compared with other state-of-the-art methods based on TEP benchmark dataset, and TDLN-trees achieves a fault detection rate (FDR) of 98.46%, which is the highest among the evaluated methods, confirming the superior performance of TDLN-trees in fault detection. The ablation study demonstrates the significance of each component of TDLN-trees in enhancing detection capabilities. The experimental outcomes reveal the promising application of TDLN-trees for handling large-scale, high-dimensional, nonlinear data.

The rest of the paper is organized as follows: section 2 reviews the basic theory of BLSTM, LSTM, and ET. Section 3 introduces the data extraction and preprocessing operations required for fault detection by TDLN-trees. Section 4 presents the structural components of TDLN-trees and the fault detection steps. An experimental analysis based on the Tennessee Eastman process is conducted in Section 5 to validate the effectiveness of TDLN-trees. Finally, Section 6 provides conclusions.

## 2 PRELIMINARIES

TDLN-trees combine the advantages of algorithms such as LSTM, BLSTM, and ET, and the following is a brief description of these methods:

### 2.1 LSTM

LSTM is a modified Recurrent Neural Network (RNN) model, as shown in Figure 1(A and B) demonstrates the network structure of RNN and LSTM. LSTM achieves the adaptive memory of important information and accurate forgetting of redundant information by adding three gating structures, namely, the forget gate $f$, the input gate $i$, and the output gate $o$. Moreover,

LSTM overcomes the problems of gradient explosion and vanishing that RNNs face when processing complex time series data[30,31].

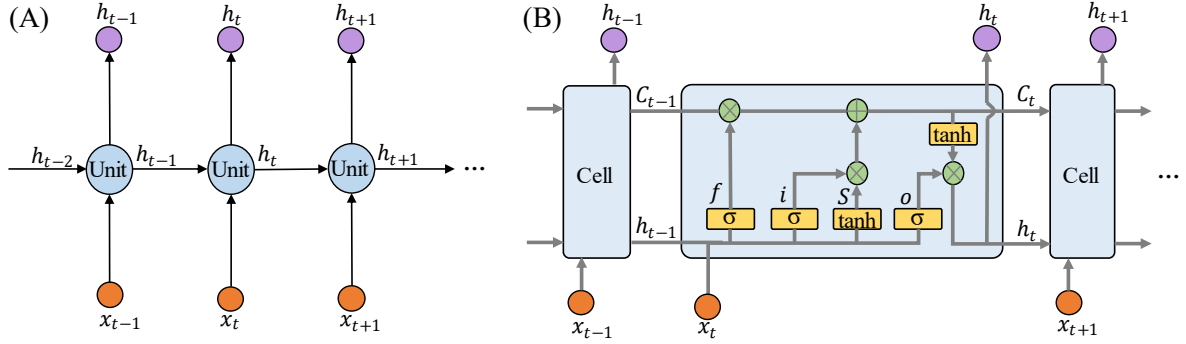

**Figure 1.** The structure of RNN and LSTM.

The forget gate $f$ determines which information from the previous time step is discarded at time t. The value of the forget gate $f$ is denoted as $f_t$ (Equation (1)). The input gate $i$ and its accompanying candidate cell state gate $S$ determine which information will be added to the memory cell at time t, denoting the values of the input gate $i$ and the candidate cell state gate $S$ as $i_t$ and $S_t$ (Equations (2) and (3)), respectively. The updated value $C_t$ of the memory cell at time t is based on the forget gate, the input gate, and the candidate state gate (Equation (4)). The output gate $o$ determines which information is output by this memory cell at time t, the value of the output gate $o$ is denoted as $o_t$ (Equation (5)), and the output of the memory cell is denoted as $h_t$ (Equation (6)). At the initial time (t = 0), $i_t$, $C_t$, and $o_t$ are typically set to a value close to 0, while $f_t$ is usually initialized to a value close to 1. The state values of each gate at time t are as follows:

$$f_t = \sigma(W_f \cdot [x_t, h_{t-1}] + b_f) \tag{1}$$

$$i_t = \sigma(W_i \cdot [x_t, h_{t-1}] + b_i) \tag{2}$$

$$S_t = \tanh(W_S \cdot [x_t, h_{t-1}] + b_S) \tag{3}$$

$$C_t = f_t * C_{t-1} + i_t * S_t \tag{4}$$

$$o_t = \sigma(W_o \cdot [x_t, h_{t-1}] + b_o) \tag{5}$$

$$h_t = o_t * \tanh(C_t) \tag{6}$$

As time goes on, the feature information of the data passes sequentially through the memory cells and is added to or deleted from the memory cells through gate structures, enabling LSTM to process time-series data effectively.

## 2.2 BLSTM

BLSTM, developed from LSTM, introduces the concepts of forward and reverse temporal directions, allowing the network to consider both past and future information of the input data sequence. BLSTM captures the before-and-after temporal relationships more comprehensively[32]. Its structure is shown in Figure 2.

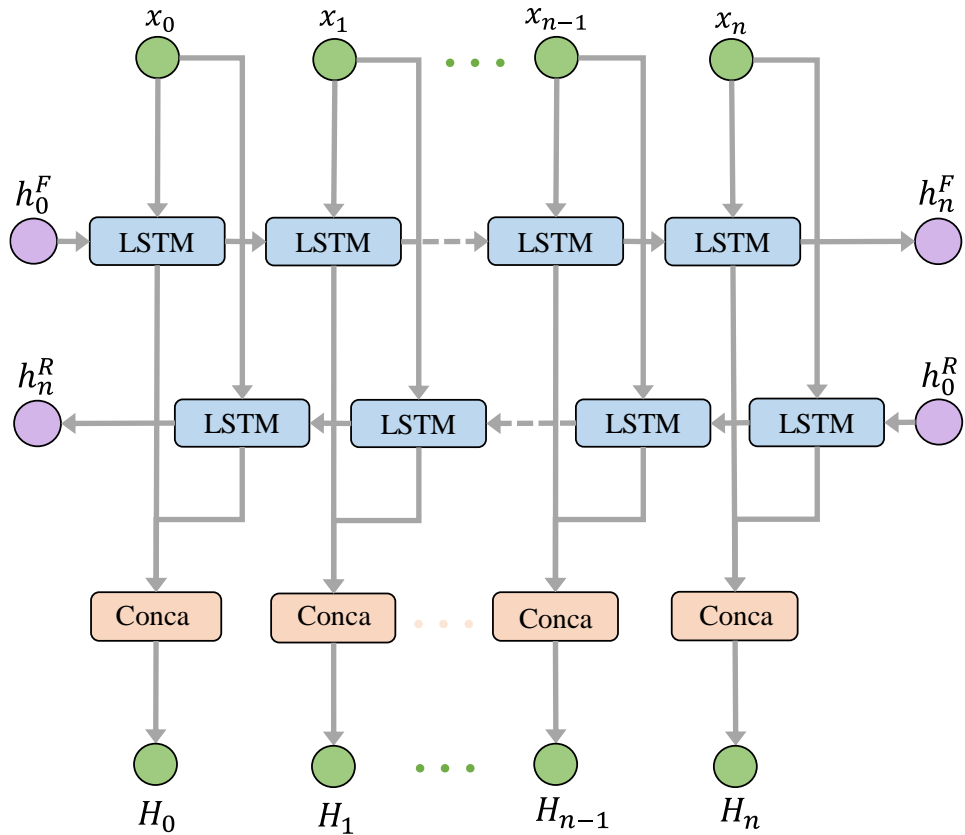

**Figure 2.** The structure of BLSTM. (Conca is the abbreviation for concatenate.)

The forward LSTM unit captures past information of the input data sequence, and the reverse LSTM unit focuses on future information. The BLSTM combines the hidden layer state $h_t^F$ of the forward LSTM unit and the hidden layer state $h_t^R$ of the reverse LSTM unit at the time t to have the output $H_t$, which are represented as follows:

$$h_t^F = LSTM(x_t, h_{t-1}^F, C_{t-1}^F) \qquad (7)$$

$$h_t^R = LSTM(x_t,\ h_{t+1}^R,\ C_{t+1}^R) \tag{8}$$

$$H_t = [h_t^F,\ h_t^R] \tag{9}$$

where $h_{t-1}^F$ and $C_{t-1}^F$ are the output values and state update values of the forward LSTM unit at the previous moment, $h_{t+1}^R$ and $C_{t+1}^R$ are the reverse LSTM unit's, respectively, [ , ] denotes the concatenating operation.

**2.3 ET**

ET, an integrated learning method, constructs unpruned decision trees in a top-down way, similar to RF, but introduces more randomness and diversity[33].

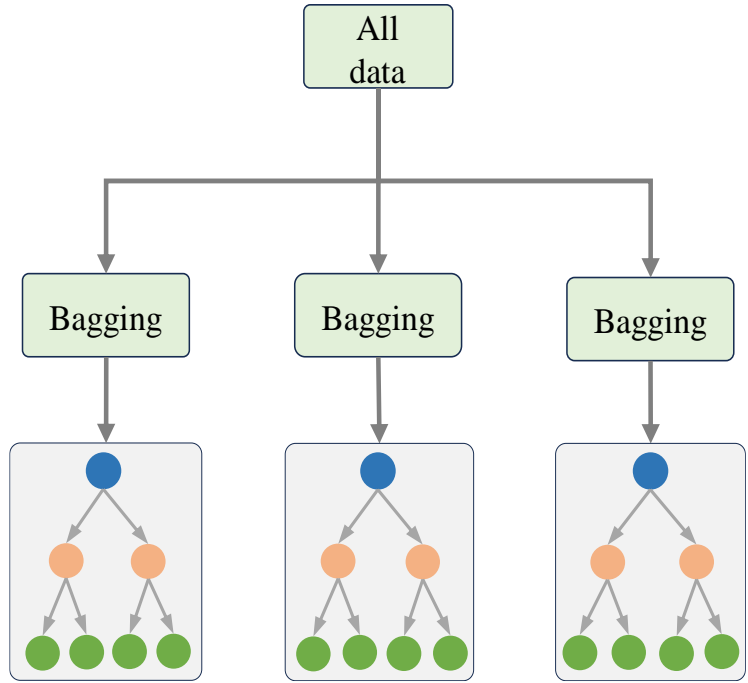

**Figure 3.** The structure of ET.

As shown in Figure 3, ET analyzes the complete raw data for constructing the decision tree. Firstly, a subset of features is randomly selected to construct a single decision tree. Secondly, for continuous feature subsets, the feature values are sorted and each unique value is considered as a potential split point; For discrete feature subsets, a potential split is considered between consecutive feature value. Finally, the Gini index for each potential division is calculated, and the division with the smallest Gini index is selected for node classification. In Figure 3, bagging represents represents bootstrap sampling on the original data, and the base layer represents the decision tree, with blue circles indicating root nodes, orange intermediate nodes, and green leaf nodes. The Gini index of the sample set D is denoted as $Gini(D)$, and the Gini index of the subset of D as $Gini_{sub}(D^v)$. These are expressed as follows:

$$Gini(D) = \sum_{v=1}^{V} \frac{|D^v|}{|D|} Gini_{sub}(D^v) \tag{10}$$

$$Gini_{sub}(D^v) = 1 - \sum_{i=1}^{|y|} (p_i^v)^2 \tag{11}$$

where $V$ denotes the total number of separable subsets of $D$, $p_i^v$ is the proportion of class $i$th samples in subset $D^v$, and $y$ is the total number of sample classes

The Gini index quantifies the influence of features on the results, and the smaller the Gini index, the more likely the samples in the node belong to the same class[34]. Therefore, ET selects the division points by calculating the Gini index, which improves the fault detection performance of the model by selecting the divisions favourable to the current node while maintaining the randomness of the tree.

## 3 PREPARATION FOR TDLN-TREES FAULT DETECTION

TDLN-trees for fault detection tasks require extracting the corresponding timing features of chemical production data and performing preprocessing operations such as normalization and one-hot encoding. The specific details are as follows:

### 3.1 Data extraction based on sliding-windows method

In this paper, we use the sliding window method to capture features with temporal correlation. The core concept of the sliding window method involves sliding a window of width $w$ and step size $s$ across the sample set until its end, thereby generating continuous and partially overlapping feature matrices while integrating fault types into label matrices. The features extracted by the sliding window method can help the model to perceive the temporal evolution of data, expressed as follows:

$$M_{F,k} = [f_{k,1}, f_{k,2}, \cdots, f_{k, \left\lfloor \frac{N-w}{s}+1 \right\rfloor}] \tag{12}$$

$$M_{L,k} = [l_{k,1}, l_{k,2}, \cdots, l_{k, \left\lfloor \frac{N-w}{s}+1 \right\rfloor}] \tag{13}$$

where $M_{F,k}$ and $M_{L,k}$ represent the feature and label matrix windows of the $k_{th}$ sample set respectively, $\left|\frac{N-w}{s}+1\right|$ is the largest integer not exceeding $\frac{N-w}{s}+1$, and $N$ is the size of sample set, requiring $k$ to be greater than $w$. In the detection stage, if $N$ is less than $w$, then $M_{F,k}$ consists of all the feature vectors of the $k_{th}$ sample set. The same applies to $M_{L,k}$.

The values of $w$ and $s$ must be finely tuned through experimental validation. The choice of $w$ should consider the temporal correlation and avoid introducing extraneous historical data, thereby lessening the computational load of model. Similarly, selecting $s$ should strike a balance between preserving adequate information and minimizing redundancy. Properly selected $w$ and $s$ can ensure the data features contain both the dynamic and static aspects of the chemical production process, facilitating subsequent processing.

### 3.2 Data normalization using data from normal operating state

Data features extracted using the sliding window method vary significantly across physical units and value ranges, necessitating to be normalized. This ensures a consistent scale for each feature type and balances the learning of model across all features.

### 3.3 One-hot encoding for sample data labels

To adapt the sample labels for the input of model, we apply one-hot encoding to the label matrix generated by the sliding window method. Specifically: for a sample with n categories, each category is mapped to an n-dimensional binary vector, collectively forming a matrix. It can eliminate the ordinal relationship between labels, enabling the model to more effectively understand and process the fault information of sample. One-hot coding is expressed as follows:

$$Encoding_j = [0, 0, \cdots, i, \cdots, 0] \qquad (14)$$

where $j$ denotes the $j_{th}$ sample, $i$ represents the $i_{th}$ labeling category ($0 \leq i \leq n$), and all positions are 0 except for the $i_{th}$ element, which is 1.

## 4 TDLN-TREES FAULT DETECTION

This section describes in detail the structure of TDLN-trees and the process of TDLN-trees for fault detection in real chemical production.

## 4.1 Establishment of the TDLN-trees

Addressing the issues of high complexity and challenging fault detection in the chemical production process, this subsection proposes a new fault detection model named TDLN-trees. Its structure is shown in Figure 4. The specific details are as follows:

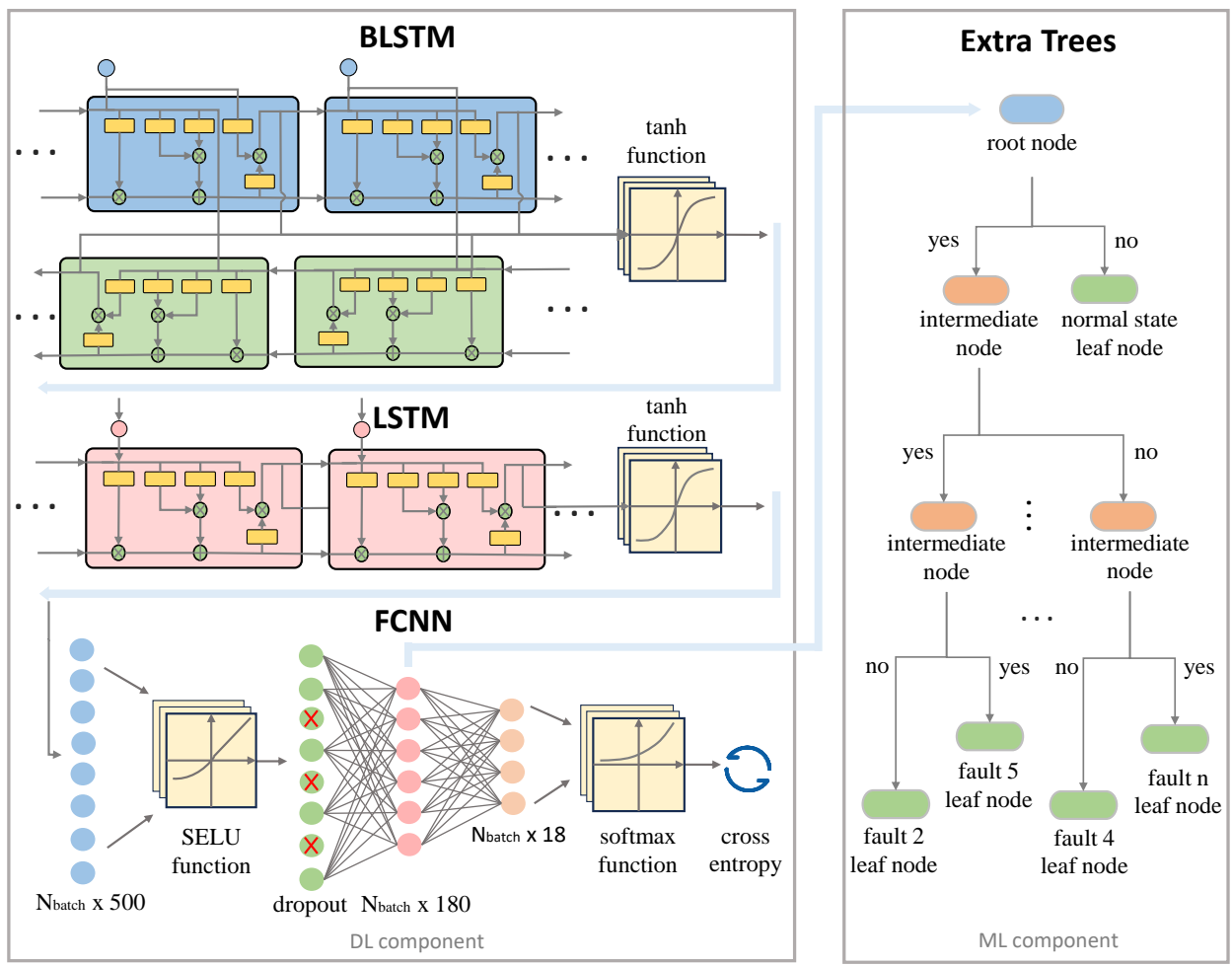

**Figure 4.** The structure of TDLN-trees.

Given the continuous nature of chemical production, process variables fall within the time series category. The BLSTM layer, constituting the first key part of TDLN-trees, is introduced initially. The addition of the BLSTM layer allows the model to capture the forward and backward temporal dynamics of industrial process variable data, its output is represented by Equation (9). Subsequently, the output of the BLSTM layer is fed into the LSTM layer, which is the second key part of TDLN-trees. This LSTM layer further improves the capacity of model to capture short-term dependencies in process variable data, with its output in Equation (6). Next, the output of the LSTM layer is relayed to the FCNN layer, the third key part of TDLN-trees. The FCNN layer can combine and transform temporal features to enhance the ability of TDLN-trees to interpret higher-level data. Its output is as follows:

$$F = \sigma (W * x + b) \tag{15}$$

$$SELU(X) = \lambda \begin{cases} X & if\ X > 0 \\ ae^X - a & if\ X \leq 0 \end{cases} \tag{16}$$

where $x$ denotes the input features, $W$ denotes the weight matrix, $b$ is the bias, the value of $\lambda$ is 1.05070098 and $\alpha$ is 1.67326324.

The BLSTM layer, LSTM layer, and FCNN layer collaboratively process the complex temporal data, constituting DL. Cross-entropy loss evaluates the fault detection capability of DL. It indicates the discrepancy between the probability distributions of the model's output and the actual labels. And we adjust the component parameters to minimize the cross-entropy loss and improve the fault detection rate of TDLN-trees.

Next is the introduction of ET, ET belongs to ML of TDLN-trees. The FCNN layer output is passed to ET, which selects division points using the Gini index (Equation (11)) and maps process variable features to corresponding fault types. Therefore, through the above steps, TDLN-trees can improve the accuracy of fault detection and reduce the occurrence of false alarms and omissions to ensure the stability and safety of chemical system operations.

**4.2 TDLN-trees fault detection process**

This section explains the application of TDLN-trees in fault detection within real chemical production. As shown in Figure 5, fault detection comprises two parts: offline fault learning stage and online fault detection stage.

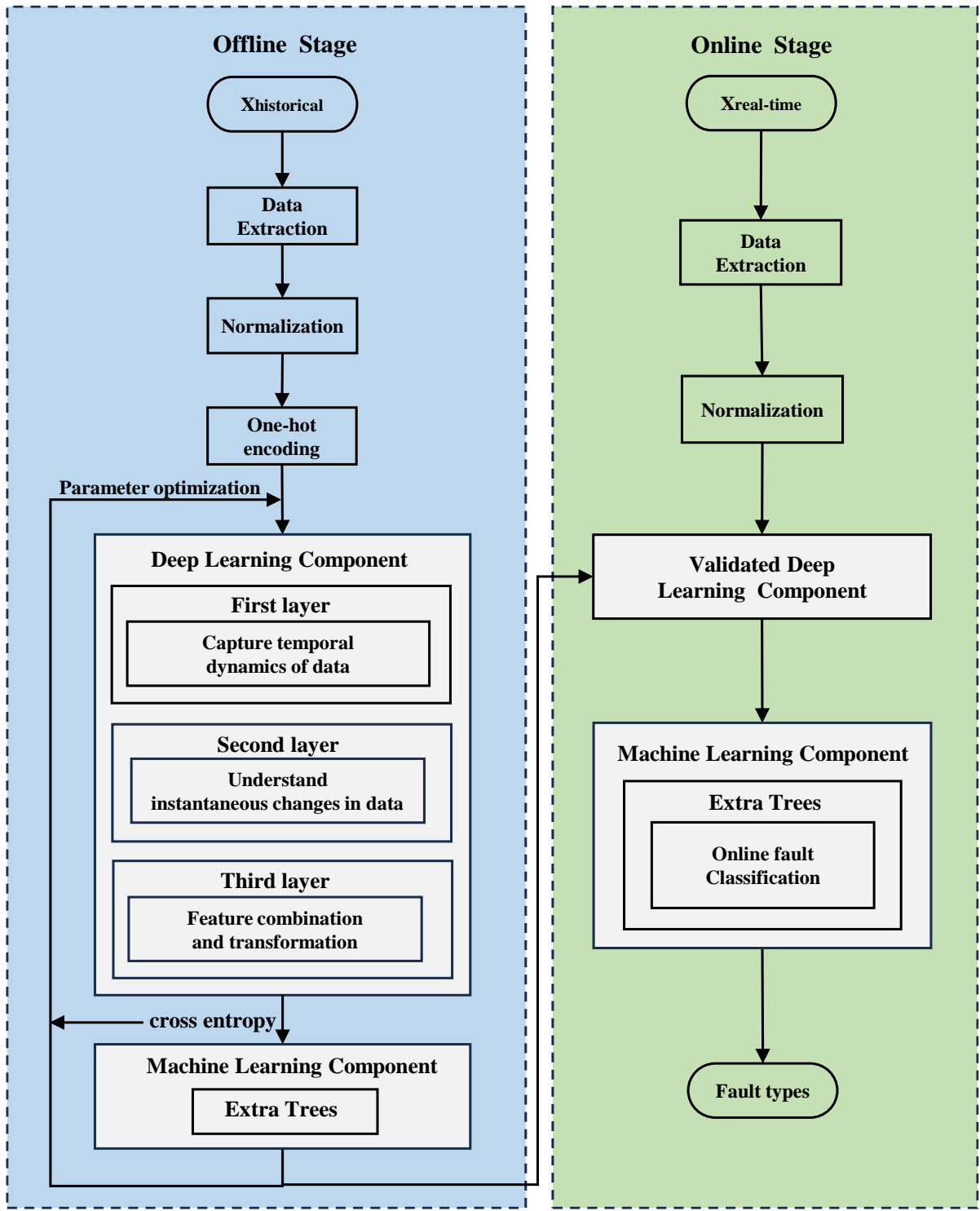

**Figure 5.** TDLN-trees fault detection process.

Offline stage fault learning: as shown in Figure 5. Firstly, the offline time series data are processed using the sliding window method to obtain the feature and label matrices with temporal correlation; Secondly, preprocessing operations are conducted on the matrix, including normalizing the fault data using normal state data, and one-hot encoding the label matrix; Then, these are divided into a training set and a validation set as inputs to the TDLN-trees, and the DL is trained, and the parameters of the DL are adjusted to compute the cross-entropy loss; Finally, input the training and validation sets into the trained TDLN-trees again, ML fits the FCNN's middle layer output of the training set, and further optimizes the parameters of DL and ML based to the classification results of ML on the output of validation set. Fault detection can be performed by realizing the above operations.

Online stage fault detection: As shown in Figure 5. Firstly, the online chemical production monitoring data is processed using the same preprocessing method described above, including sliding window extraction and normalization; Secondly, the processed data are input into TDLN-trees, and the temporal features are extracted after processing with the optimized DL; Lastly, classification is performed using ML, and Extra Trees compares the Gini indices to determine the fault classes of the monitored data.

## 5 EXPERIMENT

To validate the fault detection performance of TDLN-trees, we conduct various experiments based on the Tennessee Eastman Process (TEP) dataset, and select precision, Receiver Operating Characteristic (ROC) curves, and Fault Detection Rate (FDR) as indicators to assess the fault detection performance, precision and FDR are as follows:

$$precision = \frac{TP}{TP+FP} \quad (17)$$

$$FDR = \frac{TP}{TP+FN} \quad (18)$$

where TP represents the number of correctly detected faults; FP represents the number of incorrectly predicted faults; FN represents the number of undetected faults.

## 5.1 Tennessee Eastman Process

TEP is a complex model for simulating chemical processes, and is often utilized to test the effectiveness of process monitoring and fault detection methods[35]. As shown in Figure 6[36], the TEP model comprises five operating units: Stripping Column, Condenser, Compressor, Reactor, and Separator. For detailed roles of these operating units, refer to references [37,38]. A series of chemical reactions were carried out in these operating units, where the gas-phase reactants A, C, D, and E are converted to liquid-phase products G and H, generating by-products F, and B is an inert ingredient not participating in the chemical reaction. For the specific chemical reaction process, refer to reference [39].

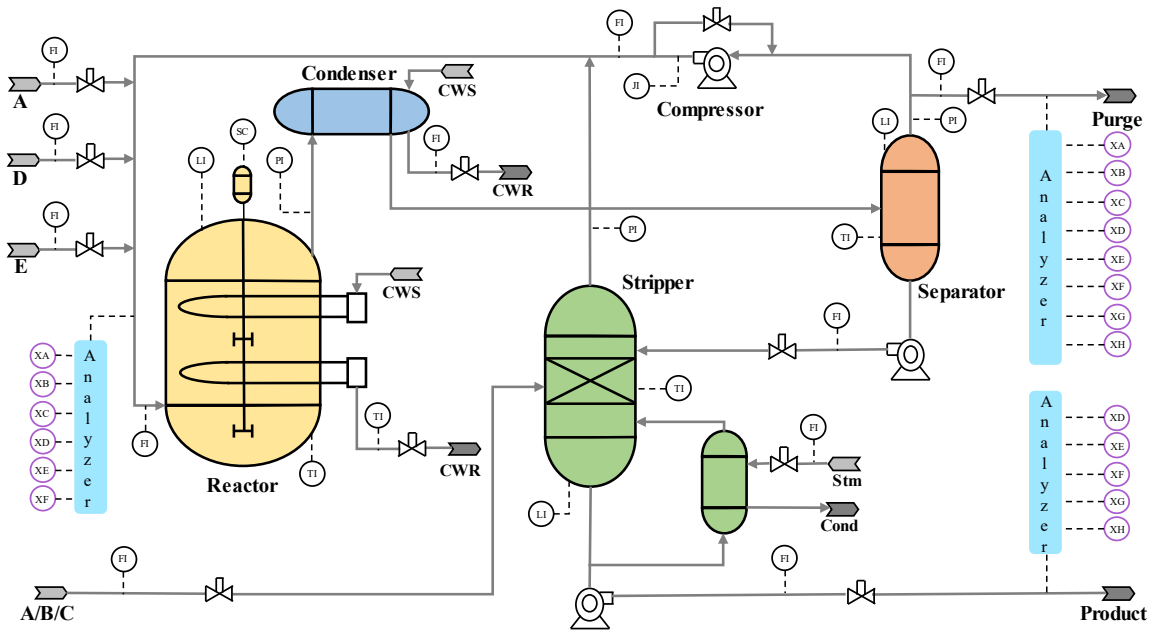

**Figure 6.** The process diagram of TEP.

TEP can simulate the normal operating state and 20 fault states of the chemical process. It contains 11 manipulated variables and 41 measurement variables, and gathering these 52 variables reflects to the operational state of the system[40]. As shown in Table 1, 0 represents the normal operating state, and 1 to 20 represents the different types of fault states. Since the mean and variance parameters did not show significant variations in fault types 3, 9, and 15, we exclude these three fault types for further analysis and evaluation. The training set consists of 500 samples for each operational state of TEP, with faults indicated after the 20th sample, over 500 simulation runs comprising 250,000 samples in total. The testing set consists of 960 samples for each operational state of TEP, with faults introduced at the 160th sample, over 500 simulation runs comprising 480,000 samples in total. Furthermore, during training, 80% of the training set is allocated for training, and the remaining 20% constitutes the validation set.

**Table 1.** TEP disturbances table.

| Fault ID | Fault description | Fault Type |
|---|---|---|
| 0 | - | normal state |
| 1 | A/C feed ratio, B composition constant (stream 4) | Step |
| 2 | B composition, A/C ratio constant (stream 4) | Step |
| 3 | D feed temperature (stream 2) | Step |
| 4 | Reactor cooling water inlet temperature | Step |
| 5 | Condenser cooling water inlet temperature | Step |
| 6 | A feed loss (stream 1) | Step |
| 7 | C header pressure loss - reduced availability (stream 4) | Step |
| 8 | A, B, C feed composition (stream 4) | Random variation |
| 9 | D feed temperature (stream 2) | Random variation |
| 10 | C feed temperature (stream 4) | Random variation |
| 11 | Reactor cooling water inlet temperature | Random variation |
| 12 | Condenser cooling water inlet temperature | Random variation |
| 13 | Reaction kinetics | Slow drift |
| 14 | Reactor cooling water valve | Sticking |
| 15 | Condenser cooling water valve | Sticking |
| 16 | Unknown | Random variation |
| 17 | Unknown | Random variation |
| 18 | Unknown | Random variation |
| 19 | Unknown | Sticking |
| 20 | Unknown | Random variation |

**5.2 Parameter Setting**

The parameters of TDLN-trees are adjusted based on the performance of the DL training and ET's furtherclassification results. After parameter tuning, if the number of BLSTM neurons is set to 600, and LSTM to 300 or more, the training and validation set accuracies (later referred to as 'accuracies') are both above 99.7%, and the training time is 601.27 s; If they are reduced to 400 and 200, respectively, the accuracies is 99.85% and 99.67%, and the time is 476.45 s; If reduced to 100 and 50 or less, respectively, the accuracies drop to at least 96%, and the time is 287.57 s. Too many neurons will lead to DL overfitting and increase training time. Conversely, the learning ability is limited. Setting these two involves a combination of theoretical and empirical considerations, this manuscript takes a compromise, as shown in Table 2. Similarly, if the number of neurons in the FCNN input layer is set to 800 or more, dropout rate is 70%, and intermediate layer is 180. The accuracy is higher than 99.68%, but the time is at least 513.78 s, and a dropout rate of 70% is too radical. If the input layer is reduced to less than 180, the dropout rate is 0%, and the intermediate layer is 0, the accuracy of training set reaches 99% or above, but the validation set is at most 95.82%, and overfitting occurs. After debugging, it is found that compared with directly reducing the number of neurons, a higher dropout rate is more beneficial for improving the model's generalization performance, and the structure of the FCCN is set as shown in Table 2. The parameter n_estimators determines the number of decision trees, and max_depth limits the maximum depth of each decision trees. Setting estimators to 200 or more and depth to at least 30, the precision of TDLN-trees can reach 98.67% or more, but the fitting time of et is at least 146.71 s. Decreasing estimators to less than 70 and increasing depth to more than 60 results in a precision of 94.53% and a time of 13.91 s. Continuing to increase the depth to more than 100 has no significant improvement on TDLN-trees. Considering that TEP is complex and DL has captured the deep temporal relationships in TEP, thus more decision trees are built to learn the data features comprehensively, as shown in Table 2. Taking the input layer as an example, 'None' represents the batch size for TDLN-trees, while 20 and 52 represent the time step and the number of features, respectively. Other parameters are set, as shown in Table 3.

Finally, we verify the effect of different window lengths w and step size s on TDLN-tress training, as shown in Table 4, and select w of 30 and s of 20 for comprehensive performance. The TEP is divided into normal state data and fault data, and segmenting it for sampling avoids the generation of mixed window and ensures no data loss.

**Table 2.** The parameters of TDLN-trees subnetwork layers.

| Layers | Component | Architecture/Parameters |
|---|---|---|
| Input | DL | None $\times$ 30 $\times$ 52 |
| BLSTM | DL | None $\times$ 30 $\times$ 256 |
| LSTM | DL | None $\times$ 128 |
| FCNN | DL | None $\times$ 500 - dropout(0.4) - None $\times$ 180 - None $\times$ 18 |
| Extra Trees | ML | n_estimators=112 max_depth=31 |

**Table 3.** Other parameter setting.

| activation function | | | batch size | optimizer |
|---|---|---|---|---|
| tanh (BLSTM) | tanh (LSTM) | selu (FCNN) | 1024 | adam |

**Table 4.** The impact of different values of w and s on training effects.

| w | s | Accuracy(%) | Time(s) |
|---|---|---|---|
| 35 | 15 | **99.67** | 598.54 |
| 35 | 20 | 99.67 | 515.08 |
| 30 | 10 | 99.66 | 675.28 |
| **30** | **20** | **99.62** | **339.57** |
| 30 | 25 | 99.46 | **321.11** |
| 25 | 10 | 99.18 | 637.36 |
| 25 | 20 | 99.04 | 335.13 |
| 20 | 5 | 98.7 | 901.97 |
| 20 | 10 | 98.5 | 621.15 |

## 5.3 Experimental Results and Analysis

5.3.1 Accuracy and loss curves

As shown in Figure 7 is the accuracy and loss curves for the training and validation set, the red line and purple line in Figure 7A represent the accuracy of the training and validation set, respectively, while the blue and green line in Figure 7B represent their cross-entropy loss. In the first 10 epochs of 50 epochs, the accuracy of the training and validation set continues to increase, and the accuracy of validation set can stably exceed 99.6% in the last 20 epochs, with the highest accuracy reaching 99.6%, corresponding to a loss of 0.0096. These results demonstrate the feasibility of TDLN-trees.

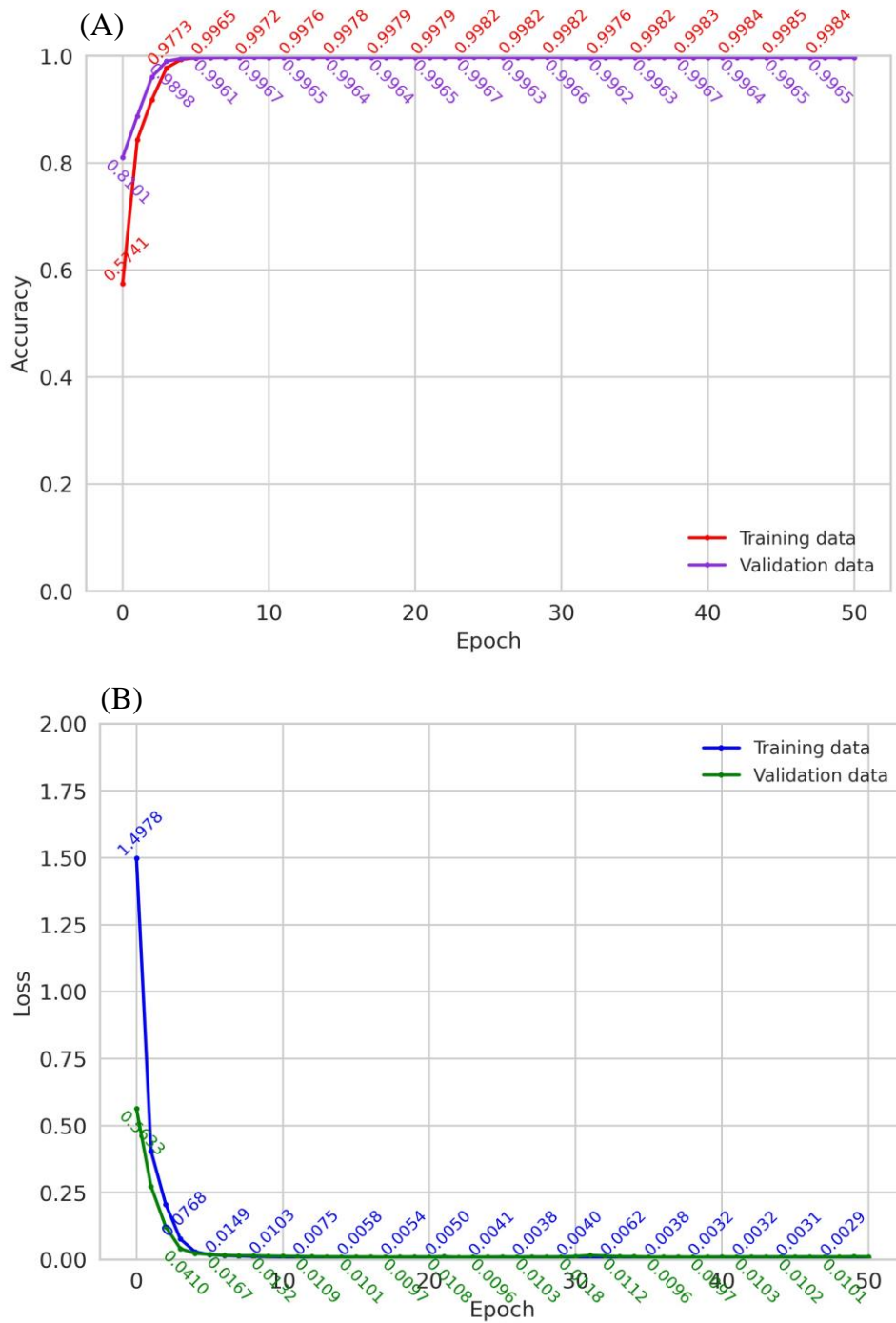

**Figure 7.** (A) The accuracy of training and validation set, (B) The loss of training and validation set.

5.3.2 Visual analysis of TDLN-trees feature learning

In order to illustrate the effectiveness of TDLN-trees after training, the feature learning capability of DL is visualized and analyzed using the t-distributed Stochastic Neighbour Embedding method (t-SNE), which can reduce high-dimensional data to two dimensions and analyze the data characteristics directly. As shown in Figure 8, Figures 8A and 8B show scatter plots of feature separability before and after processing the original data with DL, different colours indicate different fault classes. As shown in Figure 8A, most of the fault classes are mixed, whereas in Figure 8B, after DL processing, the fault classes are all distinguished into distinct clusters of different colors with minimal overlap. Therefore, TDLN-trees can fully learn the temporal characteristics of data to prepare for fault detection.

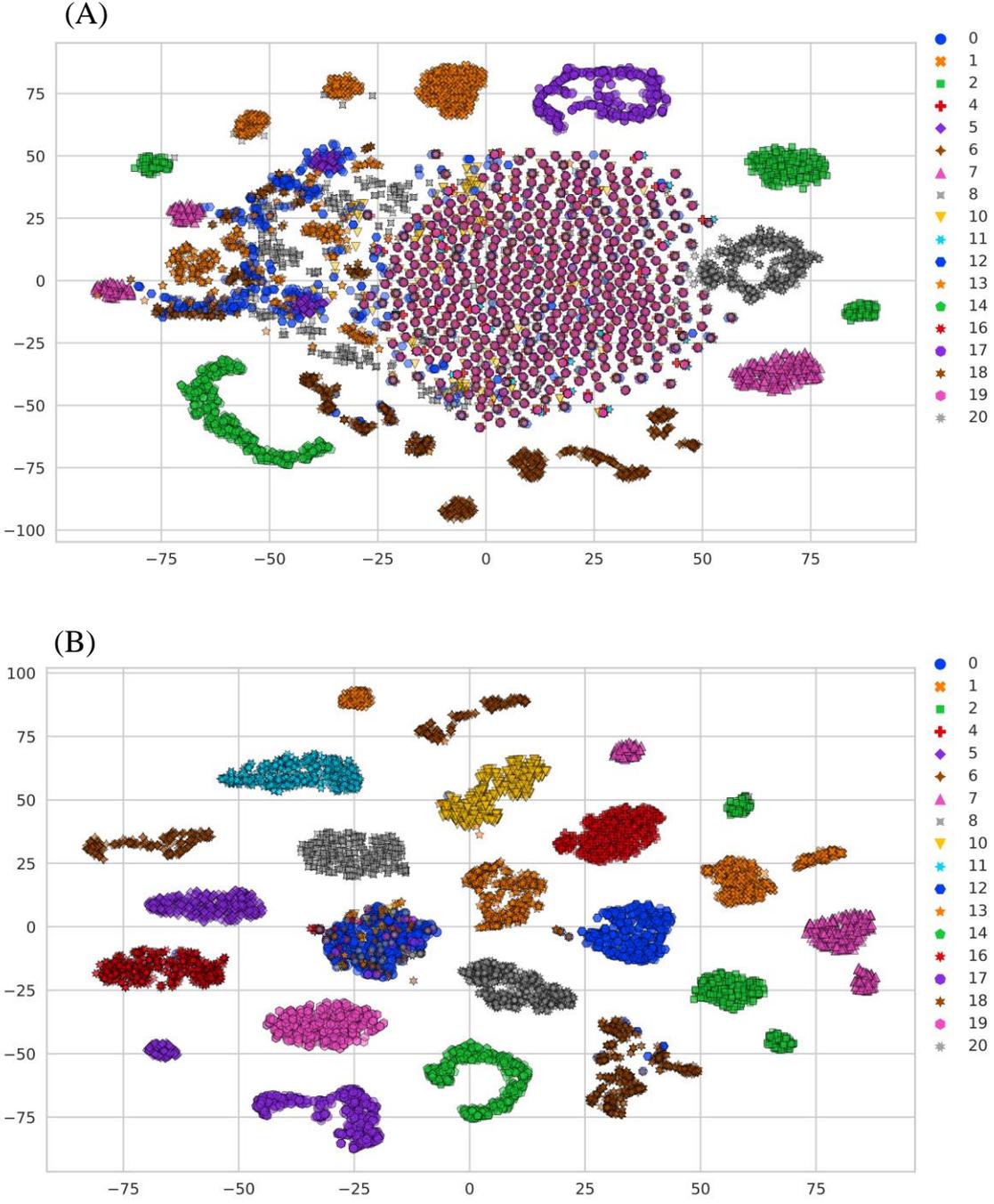

**Figure 8.** Visualization results of TEP fault detection features for t-SNE. (A) Before processing, (B) After processing

### 5.3.3 Analysis of fault detection results

To study the performance of TDLN-trees in detecting different operating states of TEP, randomly selected 2000 samples from each TEP state for testing. The fault detection results, analyzed for class prediction distribution, are shown in Figure 9, where the horizontal axis represents the actual TEP classes, and the vertical axis shows the number of samples. It is evident that the FDRs for all states exceed 90%. As shown in Figure 10, which depicts the confusion matrix for online fault detection of TDLN-trees, with rows indicating the predicted classes and columns the actual classes, the main diagonal value representing the precision of fault classification, while the off-diagonal elements indicate the proportion of misclassification. Darker colours indicate higher classification precision. Figure 10 reveals the outstanding performance of TDLN-trees in online fault detection, and the average FDR of TDLN-trees is as high as 98.46% for the 18 operational states of TEP, which confirms the excellence of TDLN-trees.

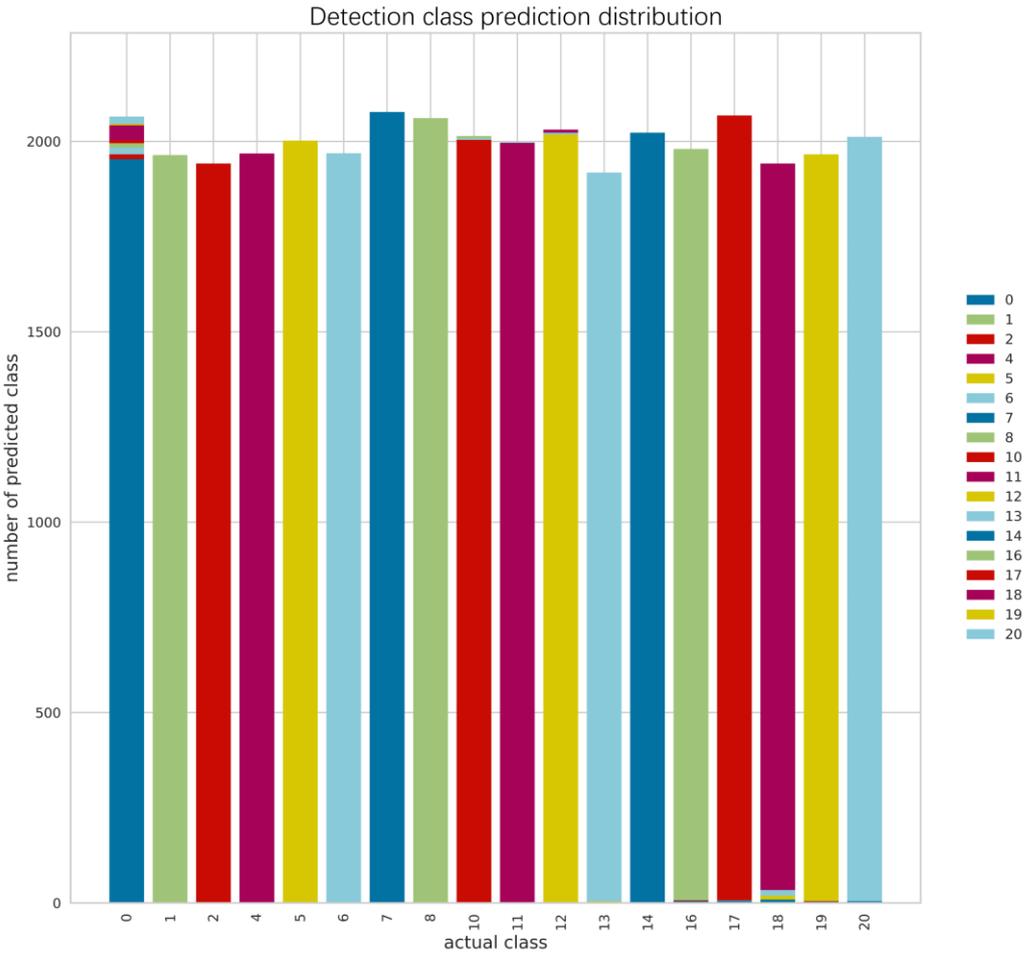

**Figure 9.** Detection class prediction distribution of TDLN-trees

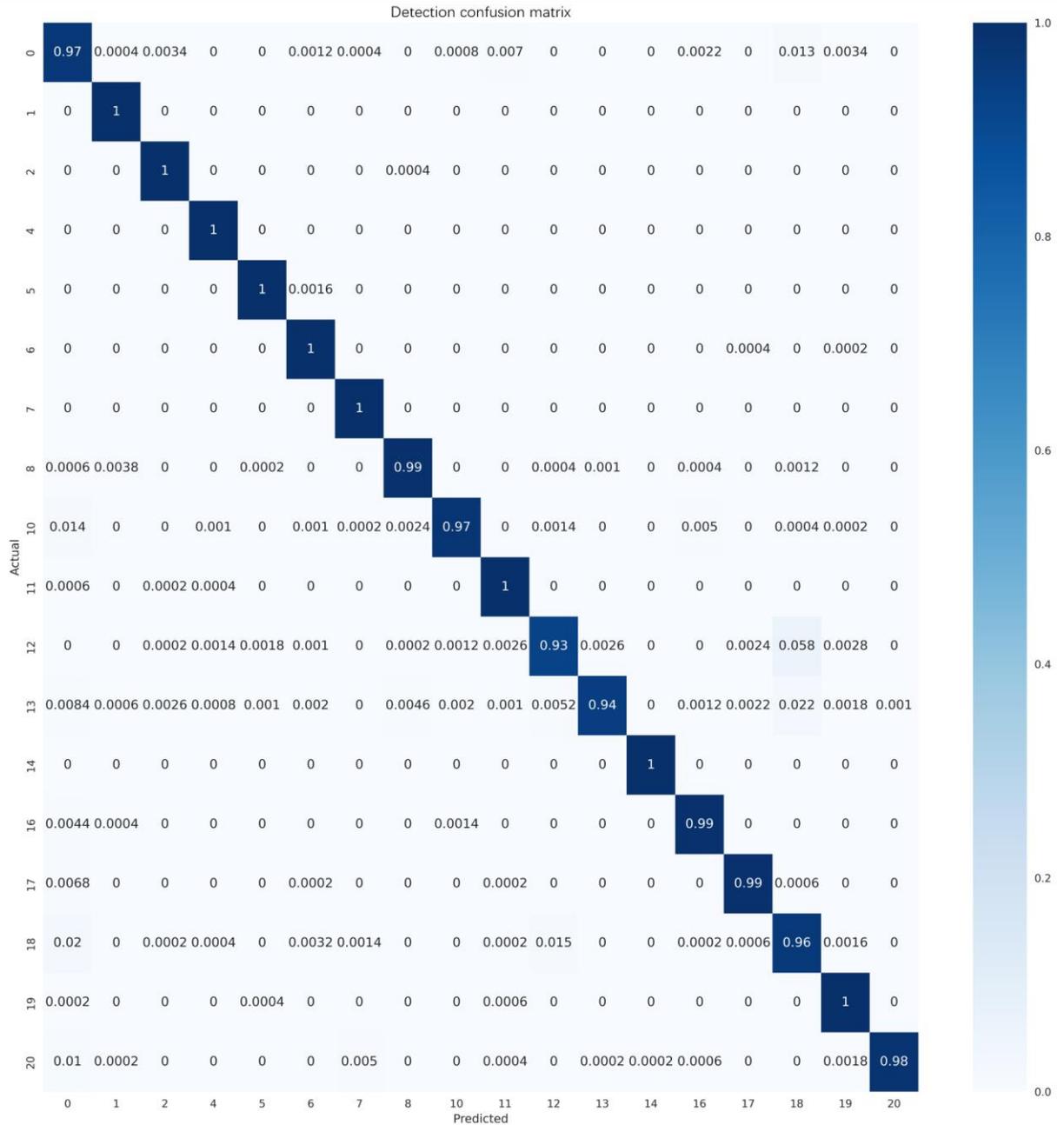

**Figure 10.** Confusion matrix of fault detection for TDLN- trees.

5.3.4 Receiver Operating Characteristic curves

The ROC curves, with the vertical axis representing the true-positive rate and the horizontal axis denoting the false-positive rate, show the trend of the true-positive rate and false-positive rate of TDLN-trees across various classification thresholds. The closer the Area Under Curve (AUC) is to 1.00, the better the TDLN-trees perform in fault detection. As shown in Figure 12, the ROC curves for the normal state and the 17 faulty states all tend to converge to the upper left of the figure, with all AUCs being at least 0.97, while the micro and macro ROC curves have AUCs of 0.99. These results demonstrate that TDLN-trees can maintain a high FDR while reducing the false-positive rate for different fault states.

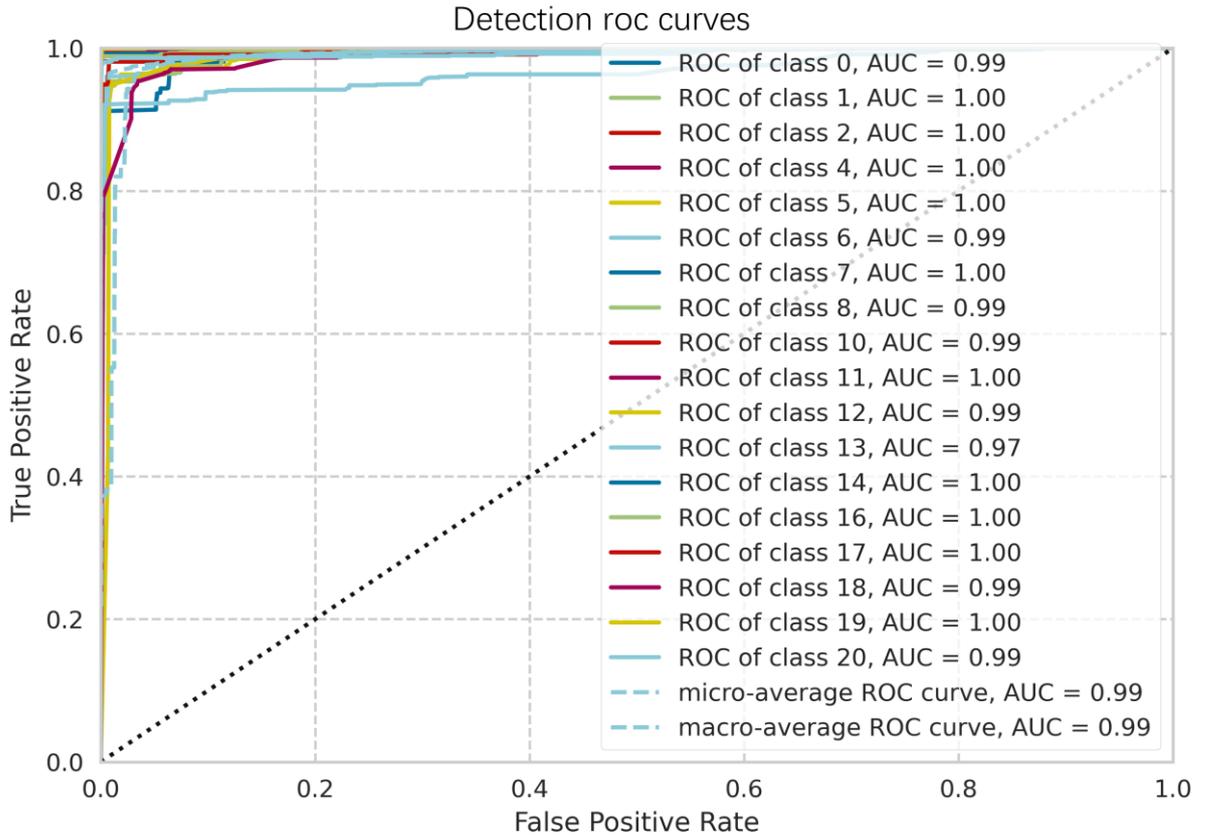

**Figure 11.** Roc curves of fault detection for TDLN-trees.

## 5.4 Comparison with other state-of-the-art methods

We compare TDLN-trees with other state-of-the-art methods for fault detection, and experimentally validate them on TEP. As shown in Table 5, the FDR of TDLN-trees is 98.56%, which is higher than that of other methods, and its FDRs for faults 1, 4, 7, and 14 are 100%. MWRSPCA[41] has an FDR of only 70.42%, its overemphasis on sparsity ignores some of the important features and affects the FDR. TceOne[42] records an FDR below 53% for faults 5 and 16, due to its One-class classifier approach that relies on features from normal state data, hindering its ability to handle fault states that are not markedly different from normal states or are highly variable. The FDR of T-BiLSTM[43] is 96.62%, and the FDR of fault 16 is less than 90%. T-BiLSTM incorporates dynamic time wrapping to consider the temporal relationship, and the fault propagation time delay will result in a small FDR in case of inconsistency with the model assumptions. DHSF-DBN[44] has an FDR 0.8% lower than TDLN-trees, and Target Transformer[45] has an FDR of 94.45%, neither matching TDLN-trees in adapting to complex chemical production data. TVAE[46] has an FDR of 2.98% lower than TDLN-trees, as it loses some original data details when compressing data into a low-dimensional space, ignoring small key features. TDLN-trees can comprehensively capture the dynamic and transient features of time series data, combining and transforming these features to form a higher level data expression with enhanced feature differentiation, and then apply ET for classifying these data to achieve improved fault detection results. This comparison experiment confirms the robustness and superiority of TDLN-trees among different faults.

**Table 5.** FDRs of TDLN-trees and other state-of-the-art methods for TEP.

| Fault | MWRSPCA | TceOne | T-BiLSTM | DHSF-DBN | Target Transformer | TVAE | Proposed method |
| --- | --- | --- | --- | --- | --- | --- | --- |
| 1 | 99.80 | 99.80 | 98.90 | 99.88 | 99.75 | 98.70 | **100.00** |
| 2 | 99.66 | 98.90 | **100.00** | 99.50 | 98.44 | 98.40 | 99.96 |
| 4 | 9.63 | **100.00** | 98.3 | **100.00** | 99.62 | **100.00** | **100.00** |
| 5 | 31.36 | 41.50 | 97.10 | **100.00** | 91.88 | 97.10 | 99.84 |
| 6 | 99.58 | **100.00** | 98.60 | **100.00** | 98.21 | 99.30 | 99.94 |
| 7 | **100.00** | **100.00** | 97.40 | **100.00** | 99.94 | **100.00** | **100.00** |
| 8 | **99.65** | 97.40 | 97.60 | 99.62 | 95.56 | 97.30 | 99.24 |
| 10 | 53.13 | 97.00 | 93.70 | 88.75 | **97.69** | 96.50 | 97.44 |
| 11 | 39.82 | 98.00 | 96.90 | 93.38 | 98.06 | 85.50 | **99.88** |
| 12 | 99.64 | 89.80 | 93.70 | **99.75** | 97.06 | 95.90 | 92.58 |
| 13 | **98.69** | 98.00 | 95.60 | 96.75 | 96.12 | 97.10 | 94.36 |
| 14 | 95.70 | 99.80 | 97.60 | **100.00** | 98.75 | 99.20 | **100.00** |
| 16 | 32.73 | 52.50 | 89.70 | 97.87 | 52.69 | 93.30 | **99.38** |
| 17 | 89.52 | 98.60 | 96.70 | 98.87 | 94.75 | 90.10 | **99.22** |
| 18 | 91.60 | 93.30 | 97.40 | 90.00 | 94.25 | **98.30** | 95.72 |
| 19 | 16.42 | 98.70 | 95.70 | 99.75 | 98.69 | 92.80 | **99.88** |
| 20 | 40.25 | 96.00 | 97.70 | 97.87 | 94.25 | 92.20 | **98.16** |
| Average | 70.42 | 91.72 | 96.62 | 97.76 | 94.45 | 95.98 | **98.56** |

### 5.5 Ablation experiment

We remove DL and ML from the TDLN-trees framework sequentially to assess their contributions, the experimental results are shown in Table 6. Compared to TDLN-tree, the FDR is reduced by 6.44% and 0.81% when DL and ML are removed, respectively, indicating that both components can improve the fault detection performance of TDLN-trees, the details described as follows:

DL: as shown in Table 6, the fault detection performance is worse when DL is removed, particularly for faults 10, 11, 16, and 20, with FDR reductions of 11.79%, 20.62%, 16.71%, and 24.35%, respectively. DL can handle the complex patterns and correlations in time series data, capture long and short-term dependencies, and refine and enhance feature representations. Thus, adding DL to the TDLN-trees framework results in better fault detection performance than without DL.

ML: as shown in Table 6, the FDRs for faults 10 and 17 of TDLN-trees are reduced by 4.06% and 3.73%, respectively, when ML is removed. ML can integrate multiple decision trees, classifying features extracted by DL using the Gini index, thereby enhancing the generalization ability of TDLN-trees across various fault types. Therefore, incorporating ML into the TDLN-trees framework can improve the FDR of TDLN-trees.

In summary, both DL and ML are crucial for enhancing the fault detection performance of TDLN-trees, with their combination offering the most significant improvement in real chemical production.

**Table 6.** FDRs of TDLN-trees in ablation experiment.

| Fault | Without DL component | Δ | Without ML component | Δ | Proposed method |
|---|---|---|---|---|---|
| 1 | 99.07 | -0.93 | 100.0 | 0 | 100.00 |
| 2 | 97.17 | -2.79 | 100.0 | 0.04 | 99.96 |
| 4 | 99.78 | -0.22 | 98.30 | -1.7 | 100.00 |
| 5 | 98.55 | -1.29 | 100.00 | 0.16 | 99.84 |
| 6 | 100.00 | 0.06 | 100.00 | 0.06 | 99.94 |
| 7 | 100.00 | 0 | 100.00 | 0 | 100.00 |
| 8 | 96.06 | -3.18 | 96.35 | -2.89 | 99.24 |
| 10 | 85.65 | -11.79 | 93.38 | -4.06 | 97.44 |
| 11 | 79.26 | -20.62 | 99.78 | -0.1 | 99.88 |
| 12 | 96.09 | 3.51 | 98.84 | 6.26 | 92.58 |
| 13 | 89.79 | -4.57 | 95.43 | 1.07 | 94.36 |
| 14 | 96.58 | -3.42 | 100.00 | 0 | 100.00 |
| 16 | 82.67 | -16.71 | 96.39 | -2.99 | 99.38 |
| 17 | 89.68 | -9.54 | 95.49 | -3.73 | 99.22 |
| 18 | 91.18 | -4.54 | 92.94 | -2.78 | 95.72 |
| 19 | 90.76 | -9.12 | 99.77 | -0.11 | 99.88 |
| 20 | 73.81 | -24.35 | 95.18 | -2.98 | 98.16 |
| Average | 92.12 | **-6.44** | 97.76 | **-0.81** | 98.56 |

# 6 CONCLUSION

In this paper, a new fault detection model named TDLN-trees is proposed for the chemical production process. It integrates the strengths of deep learning and machine learning techniques, and combines the advantages of BLSTM, LSTM, FCNN, and ET. First, the BLSTM layer in DL comprehensively analyzes the dynamic characteristics of time series data, the LSTM layer precisely identifies and captures the instantaneous changes in the data, while the FCNN layer enhances the understanding of higher-level data. Second, the ET in ML calculates the Gini index for node segmentation and realizes fault classification. In the experimental demonstration based on TEP, TDLN-trees was compared with other state-of-the-art methods and demonstrated superior performance, achieving a 98.56% FDR, surpassing that of the other methods. Ablation experiments also confirm the effectiveness of the proposed method for fault detection in chemical production. TDLN-trees have been demonstrated for TEP, and the subsequent research will involve integrating it into the mineral extraction process and developing an interpretable model that can elucidate the causes of faults.

## ACKNOWLEDGEMENTS

This work is supported by the National Natural Science Foundation of China (62203164, 62373144), Scientific Research Fund of Hunan Provincial Education Department (Outstanding Young Project) (21B0499), Hunan Provincial Department of Education (Project No. 22A0349).